# CYPUR-NN: Crop Yield Prediction Using Regression and Neural Networks

Sandesh Ramesh, Anirudh Hebbar, Varun Yadav, Thulasiram Gunta, Balachandra A

Department of Information Science and Engineering
Nitte Meenakshi Institute of Technology
Bangalore-560064

***Abstract-***Our recent study using historic data of paddy yield and associated conditions include humidity, luminescence, and temperature. By incorporating regression models and neural networks (NN), one can produce highly satisfactory forecasting of paddy yield. Simulations indicate that our model can predict paddy yield with high accuracy while concurrently detecting diseases that may exist and are oblivious to the human eye. Crop Yield Prediction Using Regression and Neural Networks (CYPUR-NN) is developed here as a system that will facilitate agriculturists and farmers to predict yield from a picture or by entering values via a web interface. CYPUR-NN has been tested on stock images and the experimental results are promising.

## I. INTRODUCTION

For centuries, agriculture is considered to be the main and the primary culture practiced all around the globe. People in the olden days have cultivated crops in their land and hence have been accommodated to their needs [1]. Predicting the yield of the crop is a vital agricultural problem. Every single farmer constantly tries to estimate how much yield can be expected from their fields. In the past, the prediction of yield was calculated by analyzing the farmer's previous results on a particular crop. Crop yield is primarily dependent on weather conditions, pests, and the planning of harvest operation. Accurate information about the history of crop yield is a vital criterion for making decisions related to agricultural risk management. The proposed method uses Regression and Neural Network techniques to predict the yield of paddy. These techniques have plenty of applications. Some of them are discussed below:

• Selection of crop and prediction of the yield- To aggrandize the yield of a crop, the identification, and selection of the ideal crop play an important role. It is also dependent on other factors like temperature, humidity, luminescence, and external pressure that surround that crop.

• Weather Forecasting- Since farmers have poor access to the internet, they heavily reliant on the little, yet vital information available concerning weather reports through newspapers or just pure hope. Artificial Neural networks have been adopted extensively for this purpose. Newly developed algorithms have shown better results over previous conventional algorithms.
• Smart Irrigation System- The groundwater levels continue to deplete day-by-day and global warming has caused drastic climatic changes. As a result, various sensor-based technologies meant for smart farming that use sensors to monitor the water level, nutrient content, weather forecast reports, and soil temperature have been introduced. Objectives of the proposed model discussed in this paper are as follows:

1. Capturing a picture of the crops to determine yield.
2. Analyzing the picture to detect diseases, if present, using Neural Networks.
3. Calculating values of external conditions such as pressure, humidity, and temperature.
4. Calculating accuracy levels of the probable yield
5. Indicating solution to the disease-causing pathogen, if present.

The paper is sectionalized as follows- Section II discusses a brief overview of the work carried out by previous researchers in the domain of crop yield prediction and the requirements needed for the same. Section III illustrates and describes the framework of CYPUR-NN and Section IV tabulates the readings obtained through experimentation of the model and methodology used. Section V discusses the results. Section VI concludes the contribution of the paper with a concise summary.

## II. LITERATURE SURVEY

As per the exploration paper by Hanks, R.J. [1], the creators utilized information mining procedures to take care of the issue of yield forecast. Various information mining procedures were utilized and assessed in farming for evaluating what's to come year's yield creation. Their paper additionally presents a concise investigation of harvest yield forecast utilizing Multiple Linear Regression (MLR) method and Density-based grouping strategies [2].
Alberto Gonzales-Sanchez, in their paper, thought about the prescient exactness of ML and direct relapse procedures for crop yield forecast in ten harvest datasets. Numerous straight relapse, M5-Prime relapse trees, perceptron multilayer neural systems, bolster vector relapse and K-Nearest neighbor techniques were positioned. Four precision measurements were utilized to approve the models: the root means square blunder (RMS), root-relative square mistake (RRSE), standardized mean outright mistake (MAE), and connection factor (R). Genuine information of a water system zone of Mexico was utilized for building the models. The outcomes indicated that M5-Prime and k-closest neighbor methods acquired the least normal RMSE mistakes (5.14 and 4.91), the most minimal RRSE blunders (79.46% and 79.78%), the most reduced normal MAE blunders (18.12% and 19.42%), and the most elevated normal connection factors (0.41 and 0.42). Since M5-Prime accomplished the biggest number of harvest yield models with the least blunders, it was an entirely reasonable instrument for monstrous harvest yield expectation in farming arranging [3].

In the paper proposed by Drummond, S.T., Sudduth, K.A., Joshi, A., Birrell, S.J. also, Kitchen, the creators' exploration depends on understanding the connections among yield and soil properties and topographic qualities in exactness farming. A vital initial step was to recognize procedures to dependably evaluate the connections among soil and topographic qualities and harvest yield. Stepwise various direct relapse (SMLR), projection interest relapse (PPR), and a few kinds of administered feedforward neural systems were researched trying to recognize strategies ready to relate soil properties and grain yields on a point–by–point premise inside ten individual site–years. To abstain from overfitting, assessments depended on prescient capacity utilizing a 5–crease cross-approval procedure. The neural procedures reliably outflanked both SMLR and PPR and gave insignificant forecast blunders in each site–year. A second period of the investigation included the estimation of harvest yield over different site–years by including climatological information. The ten site–long stretches of information were added with climatological factors, and expectation blunders were registered. The outcomes demonstrated that noteworthy overfitting had happened and shown that a lot bigger number of climatologically extraordinary site–years would be required in this kind of investigation [4].

Niketa Gandhi in their paper utilized neural systems to anticipate rice creation yield and research the variables influencing the rice crop yield. The parameters considered for the investigation were precipitation, least temperature, normal temperature, greatest temperature, and reference crop evapotranspiration, zone, creation, and yield for the Kharif season (June to November) for the years 1998 to 2002. A Multilayer Perceptron Neural Network was created. A cross-approval technique was utilized to approve the information. The outcomes indicated an exactness of 97.5% with an affectability of 96.3 and a particularity of 98.1 [5].

Simpson, G, in their paper he explored the utilization of a quick cerebellar model enunciation controller (CMAC) neural system for crop yield forecast. To begin with, expectation execution was assessed utilizing just month to month agro-met information (soil dampness, temperature, daylight). At that point, the improvement in forecast execution in the wake of fusing remote detecting information (Landsat TM) was estimated. The standard blunder was 5% when TM information was incorporated, versus 6% when TM was disregarded. The CMAC neural system applied for this investigation had recently been effectively applied in two comparable spaces: continuous cloud arrangement on Meteosat information and mineral recognizable proof with airborne obvious and infrared imaging spectrometer (AVIRIS) information [6].

In the paper proposed by Ji, B., Sun, Y., Yang, S. also, Wan, J., it was seen that by altering ANN parameters, for example, the learning rate and the number of concealed hubs influenced the precision of rice yield 3 expectations. Ideal learning rates were somewhere in the range of 71% and 90%. Littler informational indexes required less shrouded hubs and lower learning rates in model improvement. ANN models reliably delivered more precise yield forecasts than relapse models [7].

## III. METHODOLOGY

Functional requirements are those that indicate what the CYPUR-NN system will require to ensure delivery or operation. For this project, it was vital to accumulate certain requirements that will be required to attain the targets set out. A use case analysis which was implemented resulted in the following functional and non-functional requirements.

Other features of CYPUR-NN include specific criteria to judge the operation of the CYPUR-NN system. They cover ease, security, support availability, operational speed, and implementation considerations. To be more specific, the user will find it very easy to capture photos and evaluate their crops without dire effort. The system is very secure and can be easily installed. With a response time of less than 10s, CYPUR-NN is fast and reliable. One needs to consider the gravity of requirements captured from analyzing the use cases. This will aid in prioritizing the delivery of every single requirement. CYPUR-NN system was 4 designed by developing the NN model and mobile application simultaneously. The functioning of each is described in detail in the following sub-sections.

### CONVOLUTIONAL NEURAL NETWORK

One of the main functions of Convolutional Neural Networks (CNN) includes image recognition and classification. CNN image classification involves loading an image, processing it, and classifying it under categories mentioned [8]. For CNN models to train and test, each image that has been inputted will pass through serials of convolution layers with filters. Figure 1 represents the framework developed where Convolutional Neural Network architecture and Android Application were integrated to make the system in real-time.

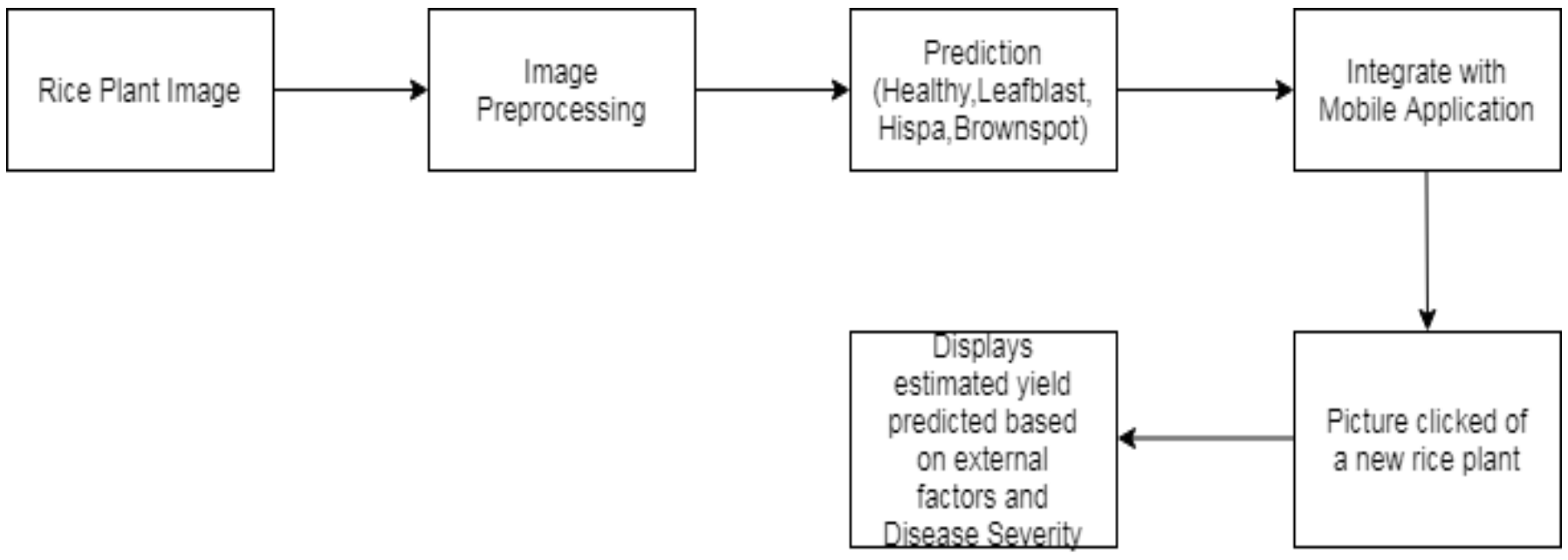

*Figure 1.Convolutional Neural Network Framework*

Figure 2 explains the working mechanism of the Convolutional Neural Network in detail. Initially, the images are collected and pre-processed to provide high-quality images for the model. In each layer, the CNN model extracts certain important features to understand and differentiate between each class.

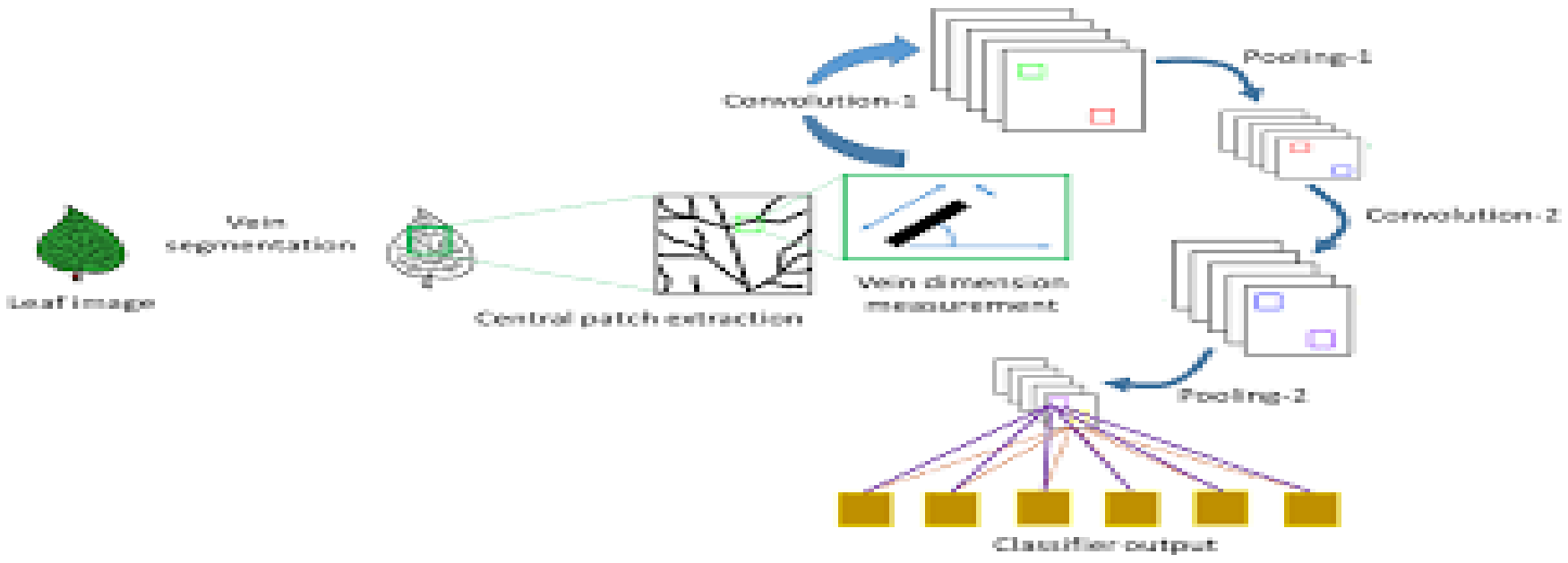

*Figure 2. CNN framework for leaf as illustrated by Gonzalez Sanchez (2014)*

Manually clicked photographs and stock images of the leaf images of the rice plant were used to train the model. The model was also trained to identify ailments that made their physical presence on the leaf and/or stem of the crop (Table 2). The accuracy metric was used to evaluate the algorithm's performance understandably. Here, the accuracy of the model was compared by the predicted label and the truth label. Categorical cross Entropy was the loss function used to optimize the algorithm. It indicates as to how poorly or how conveniently a model behaves after each step of optimization. For the hidden layers ReLU activation function (Equation 1) was used and for the output layer SoftMax activation function (Equation 2) was used. The optimizer used for the model was Adam.

$$F(x) = max\ (0, x)$$

*Equation 1. ReLU activation function*

$$P\ (y = j \mid \theta^{(i)}) = e^{\theta\ (i)} / \sum^{k}_{j\ =\ 0} e^{\theta\ (i)}{}_{k}$$

*Equation 2. SoftMax Activation Function*

## NEURAL NETWORK BASED MOBILE APPLICATION

The neural network application takes into consideration the same parameters as that of CNN. However, to predict the yield of the crop more accurately, the application also uses sensors that measure humidity, pressure, and luminescence. To test this, paddy crops were monitored in controlled conditions. The purpose of this is being, able to determine the optimum conditions for maximum crop yield. The application was also tested by replicating the conditions found in different paddy grown regions of the country. While crop yield was being measured, ailments in the crop were also being monitored simultaneously. Figure 3 shows the application displaying the possibility of the disease in the rice plant.

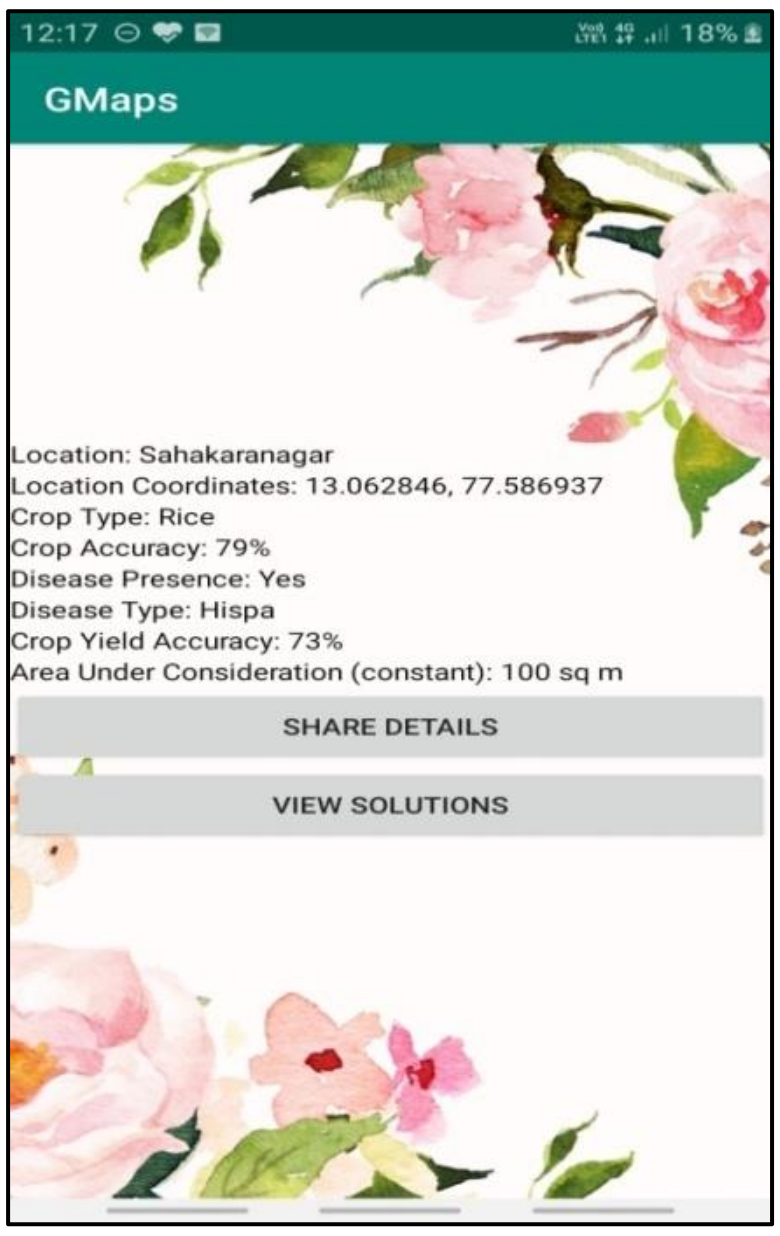

*Figure 3. Application sensors retrieving external readings*

Table 1. Contains the experimental readings of simulation conditions considered optimal and sub-optimal for the production of rice crops. For experimentation, the ph. levels of simulation soil for Punjab, Tamil Nadu, West Bengal, Andhra Pradesh, Bihar, and Karnataka were set to 7.8, 6, 6.5, 7, 8.4, and 5.5 respectively. These levels coincide with the data recorded by respective governments in their rice grown regions of the state. The area considered is constant for all readings, i.e. 100 m2. The crops were submerged in water at all times. Temperature is considered as an important parameter. Extreme high or near low temperatures, even for short, affects crop growth. High or near low air temperature decreases the growth of shoots and also reduces the growth of the root [9]. Humidity is important to make photosynthesis

possible. A good level of humidity surrounding the plant is even more vital than compared to most other crops because the plant can absorb a reduced amount of humidity and hence has less water evaporation than compared to most plants. As plants need to transpire, the surrounding humidity saturates the leaves with water vapor. When relative humidity levels are detected to be too high or there is a lack of air circulation, a plant cannot evaporate water or extract nutrients from the soil [10]. Barometric pressure has paramount effects on water chemistry and surrounding weather conditions. It has an impact on the amount of gas that can be allowed to dissolve in water. More gas, such as oxygen, can dissolve in water under higher pressure when compared to lower air pressure [11]. The impact parameter indicates whether or not the crop germination will be successful or not. To ease the purpose of understanding, we have set a threshold level of 50% yield expectancy. Any number below this is considered sub-optimal yield.

TABLE 1. QUANTITATIVE ATTRIBUTES OF NEURAL NETWORK APPLICATION

| State Conditions Under Simulation | Sample | Area (in sq. m) | Temperature (in degree Celsius) | Humidity | Pressure (in millibars) | Impact | Expected Yield (in %) |
|---|---|---|---|---|---|---|---|
| Punjab | Sample P1 | 100 $m^2$ | 14 | 38% | 138.24 mbar | Negative | 43% |
| | Sample P2 | 100 $m^2$ | 22 | 59% | 120.33 mbar | Positive | 77% |
| | Sample P3 | 100 $m^2$ | 25 | 71% | 114.46 mbar | Positive | 78% |
| | Sample P4 | 100 $m^2$ | 26 | 75% | 109.56 mbar | Positive | 89% |
| | Sample P5 | 100 $m^2$ | 38 | 88% | 98.97 mbar | Negative | 42% |
| Tamil Nadu | Sample TN1 | 100 $m^2$ | 28 | 78% | 114.67 mbar | Positive | 89% |
| | Sample TN2 | 100 $m^2$ | 29 | 78% | 115.78 mbar | Positive | 91% |
| | Sample TN3 | 100 $m^2$ | 33 | 80% | 99.45 mbar | Positive | 88% |
| | Sample TN4 | 100 $m^2$ | 35 | 84% | 96.66 mbar | Positive | 76% |
| | Sample TN5 | 100 $m^2$ | 43 | 88% | 92.34 mbar | Positive | 71% |
| West Bengal | Sample WB1 | 100 $m^2$ | 29 | 78% | 112.66 mbar | Positive | 88% |
| | Sample WB2 | 100 $m^2$ | 32 | 79% | 112.35 mbar | Positive | 88% |
| | Sample WB3 | 100 $m^2$ | 33 | 80% | 108.67 mbar | Positive | 91% |
| | Sample WB4 | 100 $m^2$ | 35 | 84% | 100.44 mbar | Positive | 92% |
| | Sample WB5 | 100 $m^2$ | 40 | 81% | 99.02 mbar | Positive | 87% |
| Andhra Pradesh | Sample AP1 | 100 $m^2$ | 30 | 78% | 99.65 mbar | Positive | 88% |
| | Sample AP2 | 100 $m^2$ | 31 | 80% | 99.78 mbar | Positive | 87% |
| | Sample AP3 | 100 $m^2$ | 35 | 82% | 91.45 mbar | Positive | 83% |
| | Sample AP4 | 100 $m^2$ | 38 | 82% | 90.89 mbar | Positive | 74% |
| | Sample AP5 | 100 $m^2$ | 43 | 87% | 84.23 mbar | Negative | 49% |
| Bihar | Sample B1 | 100 $m^2$ | 28 | 77% | 116.87 mbar | Positive | 82% |
| | Sample B2 | 100 $m^2$ | 29 | 77% | 116.72 mbar | Positive | 85% |
| | Sample B3 | 100 $m^2$ | 32 | 78% | 115.45 mbar | Positive | 85% |
| | Sample B4 | 100 $m^2$ | 33 | 78% | 115.98 mbar | Positive | 87% |
| | Sample B5 | 100 $m^2$ | 36 | 78% | 115.67 mbar | Positive | 88% |
| Karnataka | Sample K1 | 100 $m^2$ | 23 | 61% | 109.76 mbar | Positive | 79% |
| | Sample K2 | 100 $m^2$ | 24 | 61% | 109.78 mbar | Positive | 77% |
| | Sample K3 | 100 $m^2$ | 28 | 66% | 101.23 mbar | Positive | 79% |
| | Sample K4 | 100 $m^2$ | 33 | 75% | 99.78 mbar | Positive | 73% |
| | Sample K5 | 100 $m^2$ | 35 | 77% | 90.87 mbar | Positive | 71% |

Table 2 explains the accuracy of the prediction made by the model for each class i.e., Healthy, Hispa, Leafblast, and Brownspot.

TABLE 2.  QUANTITATIVE ATTRIBUTES OF NEURAL NETWORK APPLICATION

| Captured Picture | Category | Species | Ailment Condition | Accuracy |
|---|---|---|---|---|
| 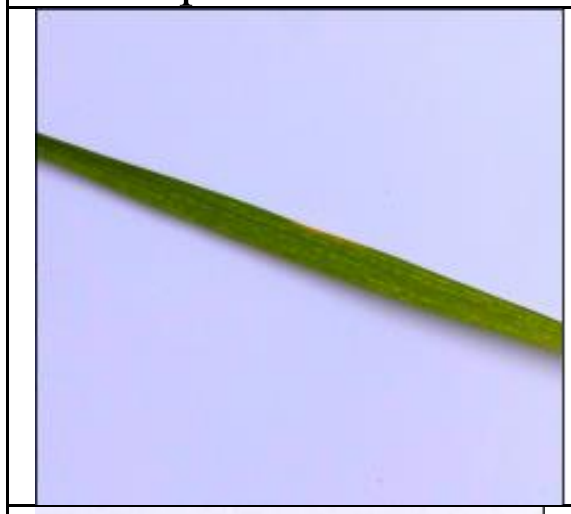 | Non-leguminous plant | *Oryza sativa* | No ailment detected in given sample | 83.7858 |
| 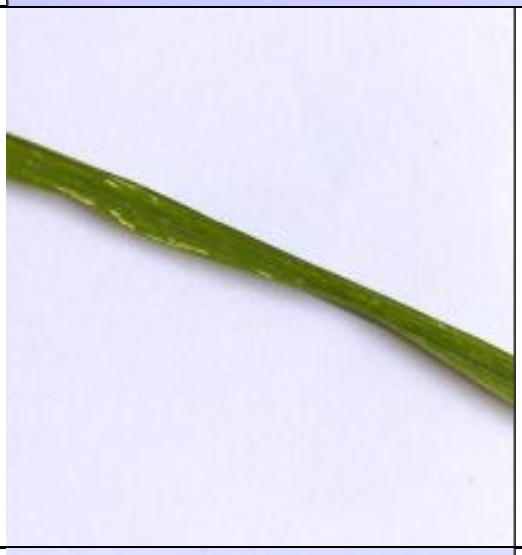 | Non-leguminous plant | *Oryza sativa* | Ailment detected in plant sample. Possible ailment-Hispa, Dryness | 84.7625 |
| 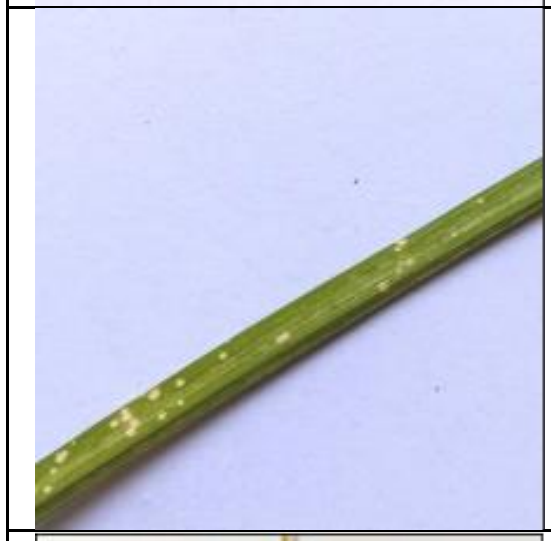 | Non-leguminous plant | *Oryza sativa* | Ailment detected in plant sample. Possible ailment-Fungal infection, Leaf Blast | 82.3554 |
| 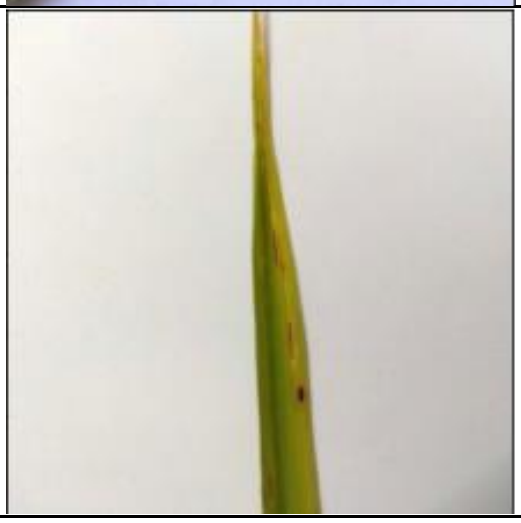 | Non-leguminous plant | *Oryza sativa* | Ailment detected in plant sample. Possible ailment-Black Spots, Brown Spots | 85.4563 |

In the next section alternative technique has been proposed to solve the early and reliable system to predict yield of the plant.

**MULTIPLE LINEAR REGRESSION**

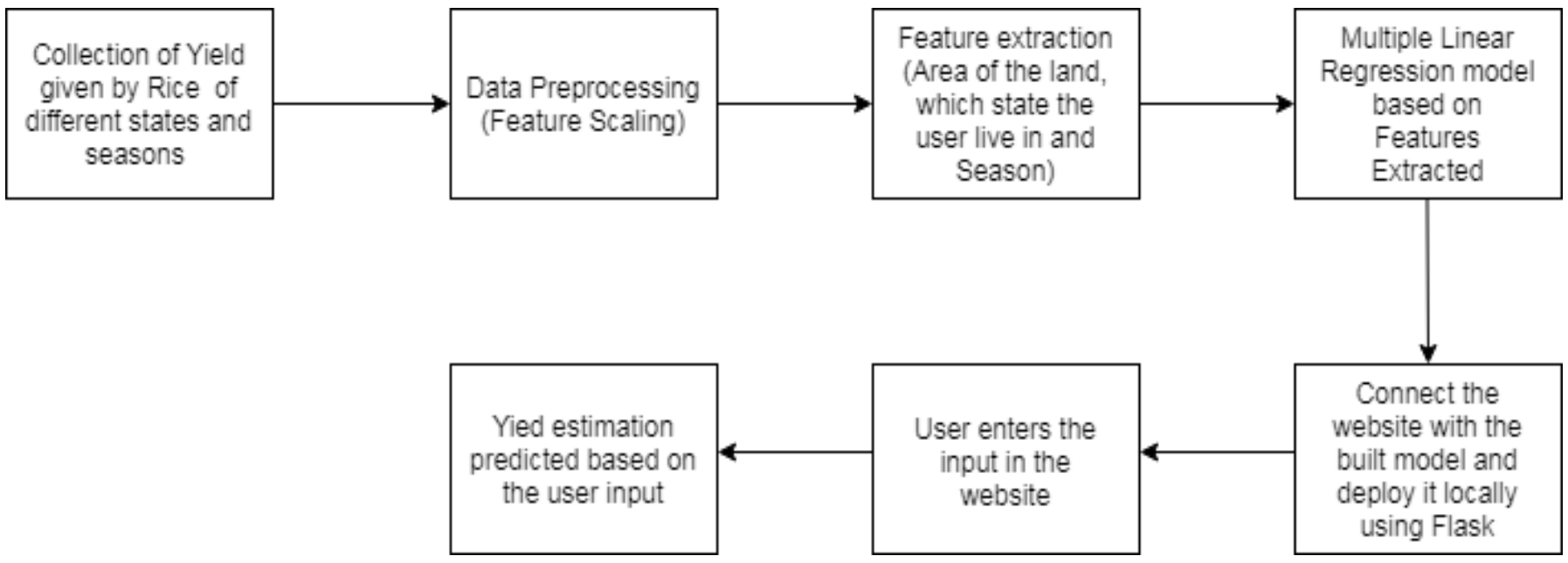

*Figure 4. Framework for Multiple Linear Regression model*

Multiple Linear Regression is a statistical technique, where multiple independent variables are used to predict one output variable. In this proposal, our 6 independent features consist of the Area of the land used by the farmer to cultivate the rice plant, the state in which he lives in and the season based on which he/she wants to know the yield of the plant. The output feature consists of the estimated yield based on the inputs given by the user. The formula of the Multiple Linear Regression is as follows:

$$y_i = b_0 + b_1 * xi_1 + b_2 * xi_2 + b_3 * xi_3 + .... + b_n * xi_n$$

*Where,*
*i = n observations,*
$y_i$ *= dependent or target variable*
$x_i$ *= independent variables*
$b_0 = y - intercept$
$b_n$ *= slope coefficients*

In this model, eleven independent features were considered since features such as state and seasons were one-hot-encoded along with the area of the land before fitting them to the model. The state feature mainly consisted of Andhra Pradesh, Karnataka, Kerala, Pondicherry, and Tamil Nadu. These features were considered keeping data distribution and curse of dimensionality in mind. The season feature mainly consisted of Autumn, Kharif, Summer, Rabi, and Winter. To package the model as a product, a simple web interface was created which will be discussed in the upcoming section.

In the Multiple Linear Regression model, we have established a relationship between the crop yield and the season. Although the size of the land does become an important factor as they are positively related to each other, we cannot expect the same yield throughout the year. Thus, keeping this in mind we have considered Kharif season as the best rice-growing season as the yield predicted in this season is higher compared to other seasons when other features such as Area and State were kept constant.

## WEB APPLICATION FOR MULTIPLE LINEAR REGRESSION

Any machine learning model is incomplete if it isn't packaged with a user interface and makes it available to the consumers. Keeping that in mind a simple web interface was designed and hosted locally using python micro-service known as Flask. Following Figure is the web interface used for the regression model (Figure.5).

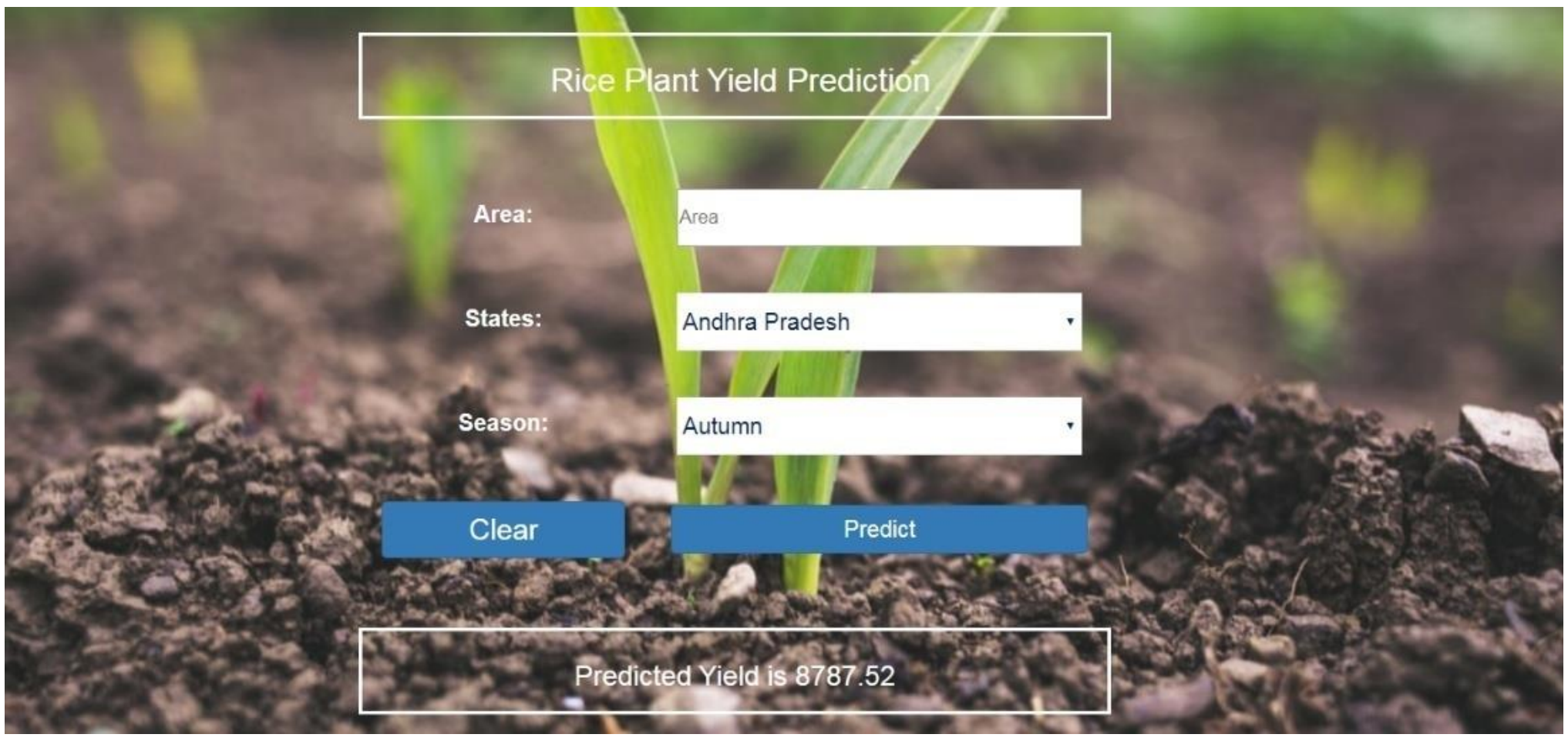

*Figure 5. Web Application for Multiple Linear Regression*

Initially, a user enters the total area in which he/she enters the total area (in hectares) in which the person wishes to grow rice plants. Next, the user is going to select the state he/she lives in and finally the season in which he wishes to choose to estimate the yield it might give. After the predict button is clicked, the model is going to consider the input values, processes it, and finally, it predicts the estimated yield which is then displayed to the user.

## IV. RESULTS AND DISCUSSIONS

### RESULTS FOR CONVOLUTIONAL NEURAL NETWORK

For the trained Convolutional Neural Network, the training set accuracy and test set accuracy were 86.37% and 83.87% respectively. The below figures shows the model performance in terms of accuracy and loss for 180 epochs.

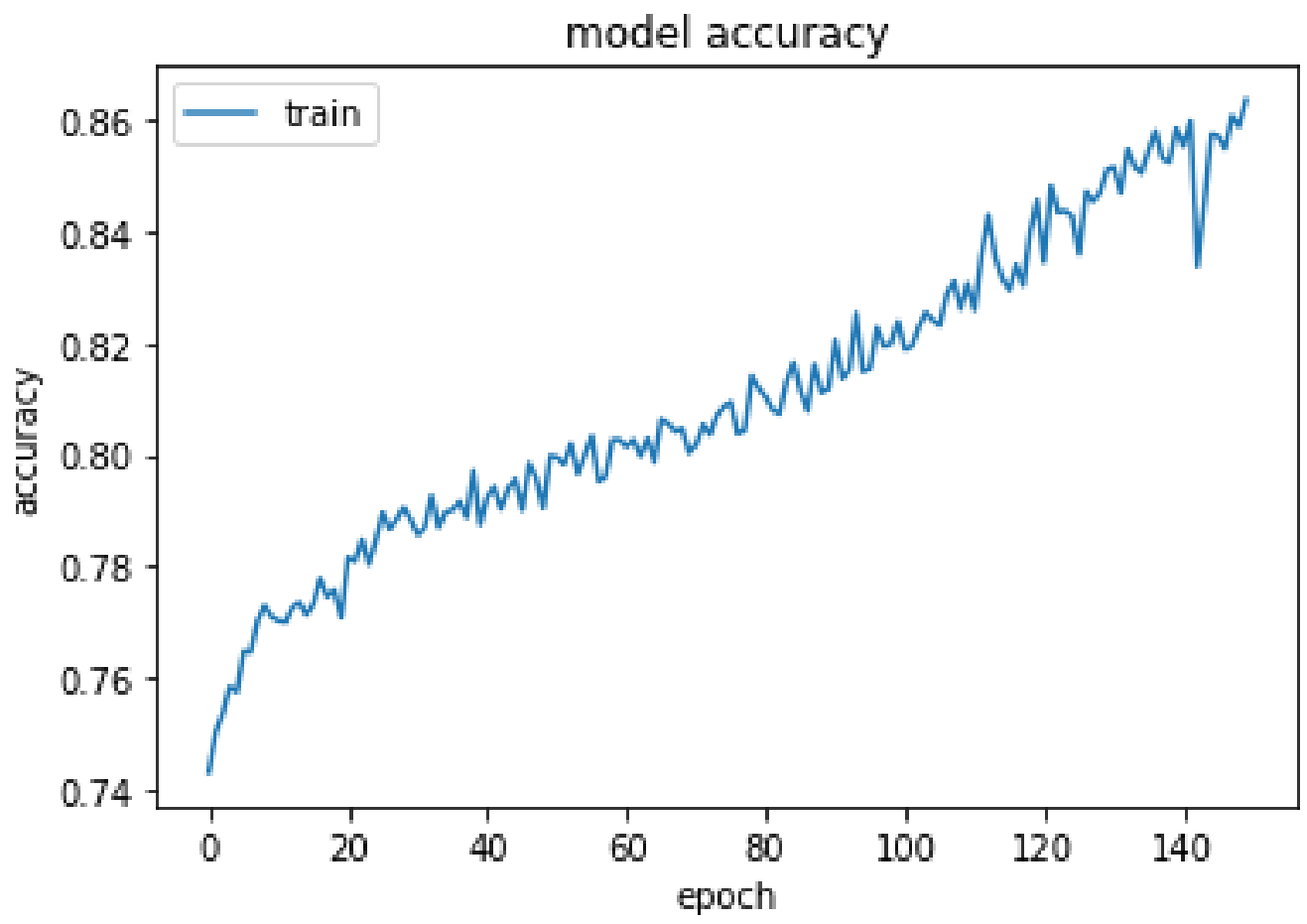

*Figure 6. Epoch v/s accuracy graph* for CNN model

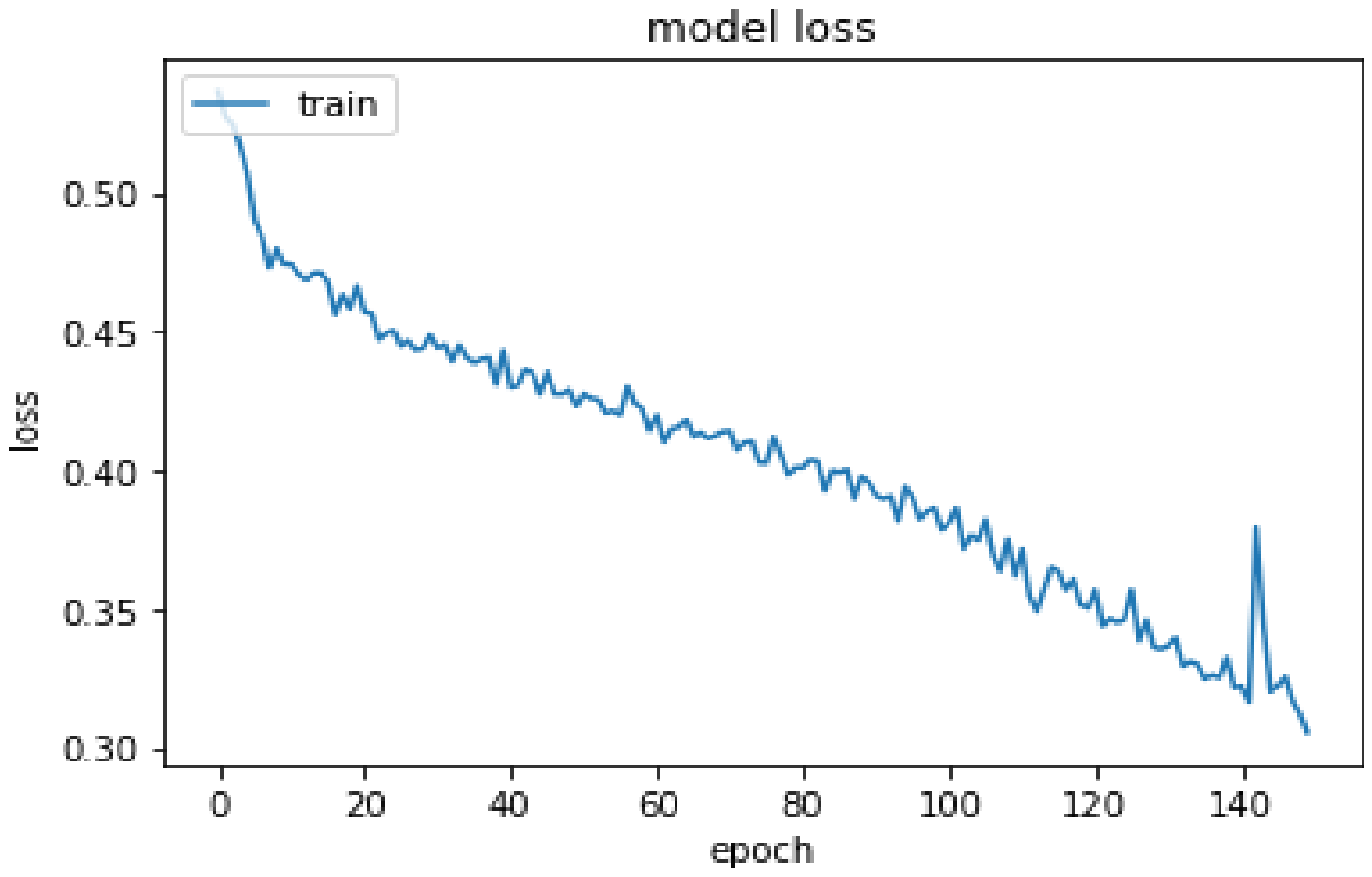

*Figure 7. Epoch v/s loss graph for CNN model*

From Figure 6 and Figure 7 we can see that as the iterations in the model increase, the accuracy of the model increases while the loss decreases accordingly, thus the model is improving its performance at each iteration. The attributes obtained by experimenting with the neural network application are tabulated in Table 1. Punjab experiences a minimum of 14°C and a maximum of 38°C. It was observed that at these two temperatures, the impact was negative and the expected yield was sub-optimal. Sample P4 with an ideal temperature of 26°C, the humidity of 75%, and an atmospheric pressure of 109.56, was expected to produce the highest yield of 89%. Tamil Nadu experiences a minimum of 28°C and a maximum of 43°C. While these conditions are considered very warm, the growth of rice remained prominent throughout changes in conditions. Sample TN2 resulted in the highest 91% yield expectancy. The corresponding conditions were 29°C, 78% humidity, and 115.78 mbar. The lowest yield expectancy was for Sample TN5, with 71%. West Bengal experiences a minimum of 29°C and a maximum of 40°C. Like Tamil Nadu, West Bengal too experiences very warm temperatures. Despite this, West Bengal has the highest rice yield compared to all the other states in the table below. With 92% yield expectancy, the conditions of WB4 are considered ideal for rice growth. Andhra Pradesh, another major rice-growing state in the country shows satisfactory yield expectancy. However, when the temperature soars to 43°C and pressure drops to 84.23 mbar, the yield too drops to a mere 49%. Bihar too had consistent results with all samples averaging between 82% and 88%. Karnataka on the other hand experiences almost moderate climatic conditions throughout the year. With the highest being 35°C and the lowest being 23°C. The optimum conditions in Karnataka are considered to be similar to Sample K3 (28°C, 66% humidity, and 101.23 mbar atmospheric pressure).

## RESULTS FOR MULTIPLE LINEAR REGRESSION

The proposal via Multiple Linear Regression is equally promising.
The Root Mean Squared value obtained for the model is 0.343. The intercept value is around 0.874.
The coefficient values for each attribute are as follows:

| Features selected for the model | Coefficients of the features selected |
|---|---|
| Area (in hectares) | 1.019 |
| Andhra Pradesh | 0.035 |
| Karnataka | -0.082 |
| Kerala | -0.309 |
| Pondicherry | 0.095 |
| Tamil Nadu | 0.260 |
| Autumn | 0.041 |
| Kharif | -0.157 |
| Rabi | -0.029 |
| Summer | 0.111 |
| Winter | 0.034 |

Based on the coefficients obtained following will be our Multiple Linear Regression Formula which can be used to estimate yield in the future.

Estimated Yield = 0.874 + 1.019*Area + 0.035* Andhra Pradesh + (-0.082)*Karnataka + (-3.09)*Kerala + 0.095*Pondicherry + 0.260*TamilNadu + 0.041*Autumn + (-0.157)*Kharif +(-0.029)*Rabi + 0.111*Summer + 0.034*Winter.

## V. **CONCLUSION**

This paper has successfully demonstrated the use of Convolutional Neural Networks and Multiple Linear Regression techniques in the development of CYPUR-NN model on a dataset consisting of various parameters related to the obtaining of expected rice yield. The model was also experimented in a simulated environment to replicate climatic conditions in various rice-growing states. The results were very realistic in nature. The agricultural sector is considered to have an important role; also, it is very important in the global economy country in the world. The use cases of machine learning have gone on to become trending, and large-scale advancements in technology have been widely utilized in modern-day agricultural technology. Artificial Intelligent techniques and methodologies are currently being used extensively in the agricultural sector as a single purpose to aggregate the accuracy and to identify solutions to the problems. As practical usage of Artificial Intelligent (AI) based on Convolutional Neural Networks (CNN) application in a plethora of fields, shows that CNN based machine learning scheme is open to change and can be implemented on an agricultural field.

## VI. REFERENCES

[1] Hanks, R.J., 1974. Model for Predicting Plant Yield as Influenced by Water Use 1. *Agronomy journal*, *66*(5), pp.660-665.

[2] Ramesh, D. and Vardhan, B.V., 2015. Analysis of crop yield prediction using data mining techniques. *International Journal of research in engineering and technology*, *4*(1), pp.47-473.

[3] González Sánchez, A., Frausto Solís, J. and Ojeda Bustamante, W., 2014. Predictive ability of machine learning methods for massive crop yield prediction.

[4] Drummond, S.T., Sudduth, K.A., Joshi, A., Birrell, S.J. and Kitchen, N.R., 2003. Statistical and neural methods for site–specific yield prediction. *Transactions of the ASAE*, *46*(1), p.5.

[5] Gandhi, N. and Armstrong, L.J., 2016, December. A review of the application of data mining techniques for decision making in agriculture. In *2016 2nd International Conference on Contemporary Computing and Informatics (IC3I)* (pp. 1-6). IEEE.

[6] Simpson, G., 1994, December. Crop yield prediction using a CMAC neural network. In *Image and Signal Processing for Remote Sensing* (Vol. 2315, pp. 160-171). International Society for Optics and Photonics.

[7] Ji, B., Sun, Y., Yang, S. and Wan, J., 2007. Artificial neural networks for rice yield prediction in mountainous regions. *The Journal of Agricultural Science*, *145*(3), pp.249-261.

[8] Guo, W.W. and Xue, H., 2014. Crop yield forecasting using artificial neural networks: A comparison between spatial and temporal models. *Mathematical Problems in Engineering*, *2014*.

[9] Panda, S.S., Ames, D.P. and Panigrahi, S., 2010. Application of vegetation indices for agricultural crop yield prediction using neural network techniques. *Remote Sensing*, *2*(3), pp.673-696.

[10] Chen, C. and McNairn, H., 2006. A neural network integrated approach for rice crop monitoring. *International Journal of Remote Sensing*, *27*(7), pp.1367-1393.

[11] Jiang, D., Yang, X., Clinton, N. and Wang, N., 2004. An artificial neural network model for estimating crop yields using remotely sensed information. *International Journal of Remote Sensing*, *25*(9), pp.1723-1732.